\documentclass[11pt]{article}

\usepackage[final]{acl}
\usepackage{algorithm}      
\usepackage{algpseudocode}  
\usepackage{amsmath}        
\usepackage{arydshln}
\usepackage{times}
\usepackage{latexsym}
\usepackage{amsmath,amssymb}
\usepackage{subcaption}

\usepackage[T1]{fontenc}

\usepackage[utf8]{inputenc}

\usepackage{microtype}

\usepackage{inconsolata}
\usepackage{multirow}   
\usepackage{graphicx}
\usepackage{xcolor}     
\usepackage{booktabs}   
\usepackage{algorithm}        
\usepackage{algpseudocode}    
\algrenewcommand\algorithmiccomment[1]{%
  \hfill{\scriptsize\color$\triangleright$~#1}%
}
\makeatletter
\newcommand{\algmarginreduce}{\setlength{\ALG@thistlm}{\parindent}}
\makeatother

\usepackage{pifont}

\title{CRAFT: A Unified Counterfactual Reasoning Framework for Tabular Question Answering and Fact Verification}

\author{
 \textbf{Chenshuo Pan\textsuperscript{1}},
 \textbf{Yu Zhao\textsuperscript{1}},
 \textbf{Jie Zhang\textsuperscript{1}},
 \textbf{Changzai Pan\textsuperscript{1}},
\\
 \textbf{Zhenhe Wu\textsuperscript{1}},
 \textbf{Jiayi Liang\textsuperscript{1}},
 \textbf{Yujie Mao\textsuperscript{1}},
\\
 \textbf{Shuangyong Song\textsuperscript{1}},
 \textbf{Yongxiang Li\textsuperscript{1}},
 \textbf{Zhongjiang He\textsuperscript{1}} 
\\
 \textsuperscript{1} Xingchen AGI Lab,China Telecom Artificial Intelligence Technology (Beijing) Co., Ltd
 \\
}

\algrenewcommand\algorithmiccomment[1]{%
  \hfill\textnormal{\normalsize\ensuremath{\triangleright}~#1}%
}

\begin{document}
\maketitle
\begin{abstract}
Table reasoning remains challenging for large language models (LLMs), particularly in tasks that require multi-step inference over long and structured tables. Existing approaches predominantly rely on single-direction reasoning, which limits their ability to explore alternative hypotheses across tasks. In this work, we propose \textbf{CRAFT}, a unified \textbf{C}ounterfactual \textbf{R}e\textbf{a}soning \textbf{F}ramework that reformulates \textbf{T}abular question answering and fact verification into a general bidirectional verification process. Our method explicitly constructs both declarative statements and their counterfactual variants. Evidence is then extracted from reasoning along both the original and counterfactual paths, and integrated via a weighted mechanism to arrive at the final answer. Experimental results show that our approach consistently surpasses representative baselines on table reasoning datasets such as WikiTQ and TabFact, achieving especially large improvements on complex question answering. Our framework also significantly mitigates performance gaps between different backbone LLMs. This indicates that counterfactual reasoning effectively overcomes the limitations of single-direction inference, guiding LLMs toward more discerning reasoning and establishing a more principled paradigm for structured reasoning tasks. Our code will be made publicly available upon acceptance.
\end{abstract}

\section{Introduction}


Tables are a representative form of structured data and are widely found in financial reports, statistical yearbooks, and various domain-specific knowledge bases \citep{liu2022tabular,chen-etal-2021-finqa}. Compared to natural language text, tables organize information explicitly through row–column structures, which poses challenges for data localization, symbolic computation, and entity linking\citep{fang2024large,WeizhengLU2025192350}. Understanding and reasoning over tables has become an important research topic in natural language processing, with table question answering (QA) and table fact verification (FV) as two representative tasks. To reason over tabular information, Large Language Models(LLMs) must interpret not only natural language questions but also the structural relations across rows and columns\citep{LIU2023100761,ruan2024languagemodelingtabulardata}. This requires the ability to perform operations such as cross-cell comparison and numerical reasoning to produce reliable inference results
\citep{badaro-etal-2023-transformers}.

\begin{figure*}[t]
  \centering
  \includegraphics[
    page=1,
    width=\textwidth,
  ]{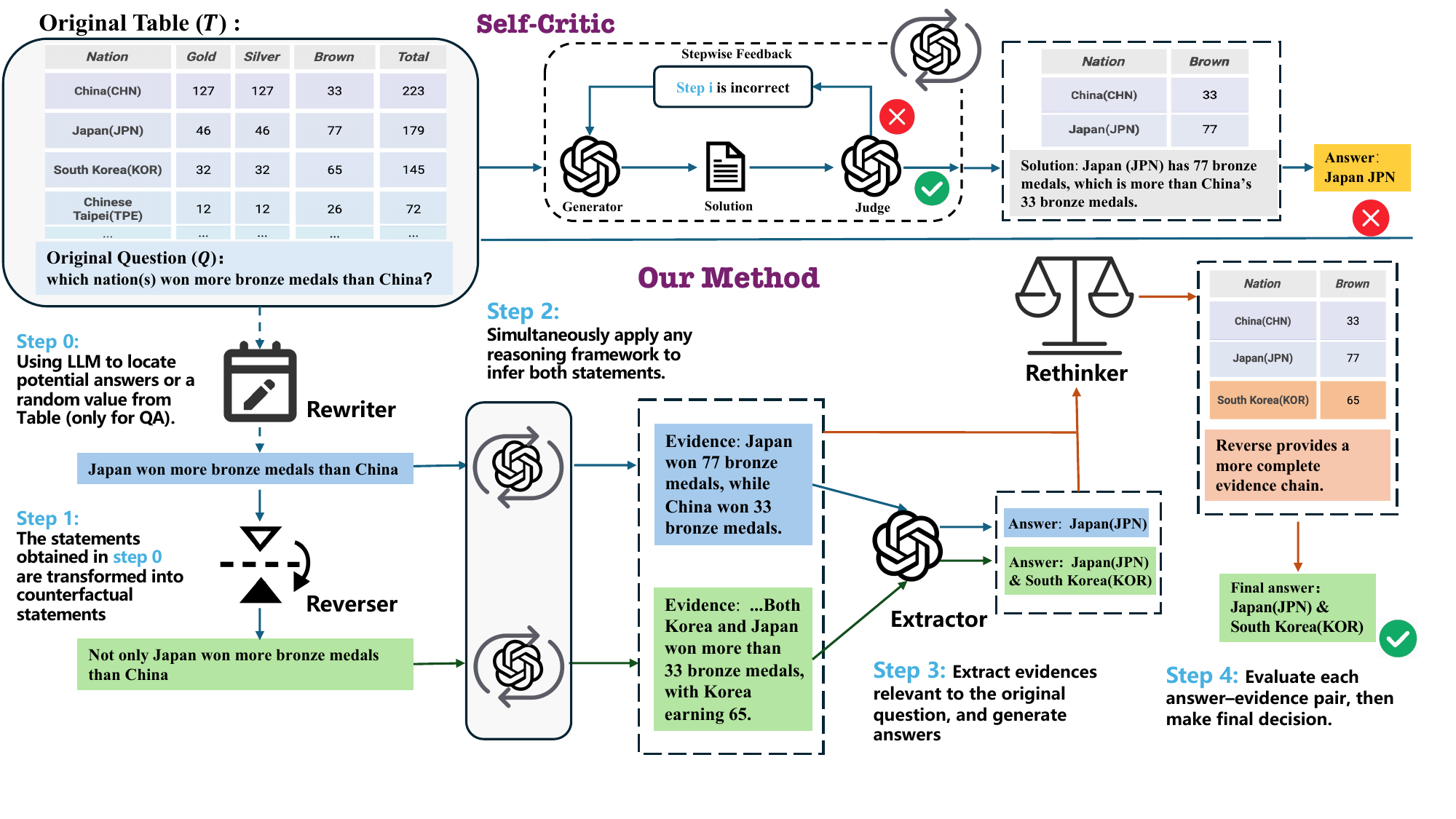}
  \caption{Overview of CRAFT, a multi-agent framework for table reasoning, Rewriter forms an initial statement from question, Reverser derives a counterfactual-inspired reverse statement to create a complementary reasoning path, Extractor gathers evidence and predictions from both paths, and Rethink jointly evaluates and generates the final answer.}
  \label{fig:my-pdf-one-third-height}
\end{figure*}




Recent work addresses tabular reasoning challenges through two main strategies. One line of research improves table understanding at the parameter level by incorporating table-specific inductive bias through pre-training or fine-tuning LLMs \citep{herzig2020tapas,su2024tablegpt2,zhang2024tablellm}, which is typically computationally expensive. Other research promote explicit reasoning and prompt engineering for sub-table extraction \citep{ye2023versatiledecomposers,sui2024tap4llm,nahid2024tabsqlify,wang2024chainoftable} to break down questions into verifiable statements without updating model parameters \cite{wei2022chainofthought,chen2023programofthoughts,Xiong2025TableZoomer}. 

Critique-based methods such as posterior self-critique \citep{yu2025tablecritic} and consistency-based aggregation \citep{ji2024treeoftable,wang2023selfconsistency,liu2024rethinkingtabular}  have recently proven effective for TableQA by detecting and correcting errors. Nevertheless, the performance gain of repeated single-direction reasoning often reduce rapidly as the number of iterations increases. This suggests that reasoning in a single direction is insufficient for robust table understanding. To effectively challenge initial premises, it is crucial to explicitly integrate counterfactual analysis into the reasoning process, which received little attention in
previous studies.

To bridge this gap, we propose \textbf{CRAFT}, a novel counterfactual reasoning framework for tabular data. CRAFT jointly constructs both supporting and counterfactual evidence chains to systematically explore alternative scenarios. Under each assumption, distinct reasoning chains are executed by LLMs, with the resulting evidence then aggregated to synthesize a final answer. This eliminates the need for extensive, task-specific pipeline engineering, offering a versatile solution for both table-based question answering and fact verification. By addressing the brittleness of unidirectional reasoning, CRAFT offers a principled approach advancing toward more structured and reliable table comprehension.

Specifically, we incorporate four cooperative modules: 1) \textbf{Rewriter} reformulates the original question into a declarative hypothesis and transforms the problem into a verifiable claim. 2) \textbf{Reverser} applies targeted transformation rules to this hypothesis to generate an informative counterfactual statement, creating a complementary scenario for reasoning. 3) \textbf{Extractor} distills essential supporting evidence from the intermediate reasoning steps of LLMs, proposing candidate answers under both factual and counterfactual assumptions. 4) \textbf{Rethinker} aggregates the extracted evidence and candidate answers to deliberate and arrive at the final, verified decision. The overall framework is illustrated in Figure~\ref{fig:my-pdf-one-third-height}. 

In summary, our contributions are three folds:

• We propose CRAFT, a counterfactual reasoning framework agnostic to specific task formulations and applies uniformly for reasoning over structured data.

• Extensive experiments and analyses are con
ducted to validate the effectiveness and robustness of CRAFT, which significantly outperforms existing methods and consistency-based strategies in both accuracy and stability.

• Our results indicate that counterfactual reasoning helps guide LLMs away from fixed patterns and toward more heuristic inference, providing a new perspective for the development of trustworthy reasoning systems.

\section{Related work}

\paragraph{Table Reasoning}

Table reasoning has gained significant attention, with many studies focusing on modifying model parameters to better adapt to the structural characteristics of tabular data. Among early attempts, TAPAS \cite{herzig2020tapas} models row–column relations through structured embeddings. A parallel line of research trains models on large collections of table-related corpora through pre-training, instruction tuning, or structured fine-tuning, exemplified by RePanda\cite{chegini-etal-2025-repanda}, enabling them to acquire tabular semantic patterns, alignment behavior, and execution abilities \cite{liu2022tapex,zhang2024tablellm,su2024tablegpt2}.Studies also incorporate reinforcement learning, using feedback signals to further enhance model performance \cite{aly-etal-2023-qa, wu2025tabler1,pan2026reasontabqacomprehensivebenchmarktable}.

On the other hand, large language models exhibit strong inherent reasoning abilities, enabling competitive performance on table tasks without modifying parameters \cite{NEURIPS2020_1457c0d6}. Prompting-based strategies have therefore been widely explored as a way to elicit this latent reasoning capability\cite{zhou2023leasttomost,chen2023programofthoughts,NEURIPS2020_1457c0d6}. Chain-of-Thought \cite{wei2022chainofthought} explicitly encourages models to articulate intermediate reasoning steps. Subsequent work extends this paradigm to tables by exploiting their structural properties. Chain-of-Table \cite{wang2024chainoftable} achieves this by breaking reasoning into executable sub-steps, where each step builds on the sub-table produced by the last. Other methods condense or reorganize table inputs prior to reasoning to ensure that models operate on compact, task-relevant data \cite{sui2024tap4llm, nahid2024tabsqlify}, or incorporate external symbolic execution frameworks that offload computation and validation to SQL or Python engines \cite{cheng2023binding,zhang2023reacTable, ni2023lever, mao2024potable}, allowing LLMs to concentrate on planning rather than calculation.

Building further, other studies emphasize multi-round and structural reasoning mechanisms for robustness and corrective capability. Examples include self-consistency \cite{wang2023selfconsistency,2025.coling-main.368}, tree-structured \cite{ji2024treeoftable} and graph-structured reasoning \cite{2025.coling-main.368} to enhance stability through multi-branch logical inference. MIX-SC \cite{liu2024rethinkingtabular} combines textual and symbolic reasoning, while Table-Critic \cite{yu2025tablecritic} introduces critical assessment to allow self-correction of intermediate steps. However, existing table-reasoning methods generally reinforce inference around the original question and do not explore the use of counterfactual reasoning paths for explicit evidence contrast.

\paragraph{Counterfactual Reasoning}

Many studies construct counterfactual variants of inputs to enhance causal stability \cite{feng2021empowering}, strengthen out-of-distribution generalization through relation-driven counterfactual contrasts \cite{yang2023relationcounterfactual}, and test inference discrimination under atypical reasoning conditions \cite{webb2025emergentanalogical}. It has also been employed for semantic delimitation and self-calibration, such as distinguishing hypothetical from factual knowledge \cite{li2023counterfactualreasoning} and improving contextual faithfulness through counterfactual exemplars in prompts \cite{zhou2023contextfaithful}. The Direct–Indirect Reasoning \cite{zhang2025indirectreasoner} framework provides a more general paradigm for solving mathematical problems by jointly incorporating forward and reverse verification. Kim introduces Counterfactual-Consistency Prompting \cite{kim2025counterfactualconsistency}, which constructs temporal counterfactual questions to enforce consistency, showing strong performance in relative temporal reasoning. However, this approach relies exclusively on counterfactual and mainly to be effective for time-related  verification tasks. To our knowledge, counterfactual reasoning has not been explored in table reasoning, even though the structured nature of tabular data makes it naturally compatible to counterfactual validation.


\section{Method}

\noindent\textbf{Notation and Definitions.}
The overall framework of CRAFT consists of four modules---\textsc{Rewriter}, \textsc{Reverser}, \textsc{Extractor}, and \textsc{Rethinker}. To formalize the problem setting, we denote the table as $T$, ground-truth answer as $A$, the natural-language question as $Q$, and the declarative statement derived from $Q$ as $S$. In table fact verification (FV), the input is already a declarative claim, and we have $S = Q$.

\subsection{Rewriter}

Rewriter module is designed to transform an open-end table QA task into an
informative and verifiable declarative statement that facilitates the
subsequent generation of counterfactual statement. Formally, the Rewriter process can be written as:
\begin{subequations}\label{eq:rewrite}
\begin{equation}\label{eq:atomic-value}
a \sim \Omega_{T,Q}
\;:=\;
\{\, v \mid \tau(v)=\tau_{A_Q} \,\}.
\end{equation}
\begin{equation}
S = \mathcal{M}(Q, a).
\end{equation}
\end{subequations}

We first infer the expected answer type $\tau_{A_Q}$ from the semantics of $Q$ alone, without refer to any ground-truth answer at inference time. Here, $\tau_{A_Q}$ denotes the semantic type expected by $Q$ (e.g., time, number, or boolean), and $\tau(v)$ denotes the semantic type of a candidate value $v$. Accordingly, $\Omega_{T,Q}$ denotes the type-consistent candidate value space induced by $T$ and $Q$. An atomic value $a$ is sampled from this space and used to instantiate the semantic slot in $Q$, yielding a declarative statement $S$. In practice, the LLM is prompted to infer $a$ from cells in $T$. When no explicit cell is available, it may randomly instantiate an $a$ with a type-consistent value grounded in $T$ and $Q$.

Unlike prior work that primarily uses QA pairs to improve fact verification~\citep{aly-etal-2023-qa}, our \textsc{Rewriter} explicitly bridges QA and FV by transforming a question into a declarative statement $S$ via type-consistent instantiation. In this way, QA is recast into the same statement-based form as FV, allowing the downstream modules to function uniformly across both tasks.

\subsection{Reverser}

The Reverser module takes a declarative statement $S$ as input and aims to construct an optimal counterfactual statement $R^{*}$. Counterfactual reasoning is not a simple binary flip; instead, a statement $S$ may admit
multiple counterfactual directions. In particular, when the $Q$ is available, $S$ may carry latent semantics beyond its surface form, so the admissible counterfactual directions should be defined with respect to the joint semantics of $(Q,S)$.
For clarity, we denote counterfactuals instantiated along different directions by
$R_i=\phi^{\delta_i}(S)$, where $\delta_i$ simply indexes distinct admissible
directions under the semantic context. Prior work has shown that models may underutilize information depending on input position or presentation  \citep{liu2024lost,wan2025positional}.Therefore, we introduce the notion of a latent reasoning space $\Psi(\cdot)$ to denote the latent reasoning space induced by a declarative statement during inference.
Under this notation, counterfactuals constructed along different directions
are characterized by inducing different reasoning subspaces:
\begin{equation}
\delta_i \neq \delta_j
\;\Rightarrow\;
\Psi(R_i)
\neq
\Psi(R_j).
\end{equation}
The purpose of introducing counterfactuals is to broaden the reasoning space beyond that induced by $S$ alone.
We characterize the optimal counterfactual $R^{*}$ as the one whose induced reasoning,
when considered together with that of $S$, provides the greatest overall
informational contribution:
\begin{equation}
R^{*}
\;=\;
\arg\max_{i}\;
\mathcal{G}\!\left(
\Psi(S)\;\cup\;\Psi(R_i)
\right).
\end{equation}

To construct an informative reverse statement $R^{*}$, we first generate a small set of candidate counterfactuals from $S$ using a rule-based template set derived from statement semantics (and the original question $Q$ when available), ensuring that the candidates remain plausible and semantically aligned with the original statement. We then prompt the model to generate a SQL-style verification program for $S$ and each reverse candidate, conditioned on the table’s column names and the entities mentioned in the statement, and use the resulting program structure as a proxy for the induced reasoning space. The final reverse statement $R^{*}$ is selected as the candidate whose induced program yields the largest structural expansion beyond that of $S$. While more advanced SQL generation techniques may further improve performance, we intentionally adopt a lightweight generation strategy in this work. More implementation details are provided in Appendix~\ref{app:reverse}.

\captionsetup[table]{position=bottom}
\begin{table*}[t]
\centering
\scriptsize
\setlength{\tabcolsep}{3pt}
\renewcommand{\arraystretch}{1.25}
\newcommand{\NA}{--}
\resizebox{\textwidth}{!}{%
\begin{tabular}{l*{10}{c}}
\toprule
\multirow{2}{*}{Method} &
\multicolumn{2}{c}{Deepseek-R1-14B} &
\multicolumn{2}{c}{Qwen2.5-72B} &
\multicolumn{2}{c}{Llama3.3-70B} &
\multicolumn{2}{c}{GPT-5-mini} &
\multicolumn{2}{c}{Average} \\
\cmidrule(lr){2-3}
\cmidrule(lr){4-5}
\cmidrule(lr){6-7}
\cmidrule(lr){8-9}
\cmidrule(lr){10-11}
& WikiTQ & TabFact & WikiTQ & TabFact & WikiTQ & TabFact & WikiTQ & TabFact & WikiTQ & TabFact \\
\midrule
End-to-End                       & 54.3 & 91.8 & 57.3 & 85.6 & 51.1 & 83.4 & 85.2 & 94.7 & 62.0 & 88.9 \\
Few-Shot                         & 70.0 & 91.3 & 60.6 & 86.0 & 62.0 & 83.1 & 85.7 & 95.1 & 69.6 & 88.9 \\
Binder \cite{cheng2023binding}      & 68.5 & 89.5 & 65.5 & 84.0 & 57.6 & 79.9 & 84.2 & 94.5 & 69.0 & 87.0 \\
Dater \citep{ye2023versatiledecomposers} & 71.6 & 92.1 & 63.8 & 89.7 & 63.3 & 89.9 & 84.9 & 95.9 & 70.9 & 91.9 \\
Chain-of-Table \cite{wang2024chainoftable} & 75.1 & 92.7 & 68.3 & 89.7 & 62.1 & 89.9 & 86.2 & \underline{96.0} & 73.0 & 92.1 \\
Critic-CoT \cite{zheng2024criticcot} & 75.2 & 92.5 & 69.0 & 89.8 & 66.8 & 88.0 & 86.1 & 94.9 & 74.3 & 91.3 \\
Alter \cite{zhang2025alter}         & 67.4 & 90.5 & 70.9 & 87.4 & 67.8 & 89.8 & \NA & 94.6 & 68.7 & 90.6 \\
Table-Critic \cite{yu2025tablecritic} &
76.3 & 93.3 &
77.7 & \underline{92.9} &
70.1 & 92.1 &
86.7 & 95.5 &
77.7 & 93.5 \\
\midrule
\textbf{CRAFT w/ Chain-of-Table Extractor} &
\underline{77.2} & \underline{93.6} &
\underline{78.7} & 92.2 &
\underline{79.9} & 92.0 &
\underline{87.0} & \textbf{96.7} &
\underline{80.7} & \underline{93.6} \\

\textbf{CRAFT} &
\textbf{79.9} & \textbf{94.2} &
\textbf{80.4} & \textbf{93.9} &
\textbf{81.4} & \textbf{94.1} &
\textbf{87.8} & \underline{96.2} &
\textbf{82.4} & \textbf{94.6} \\
\bottomrule
\end{tabular}%
}
\caption{Performance comparison between CRAFT and representative table-reasoning baselines on WikiTQ and TabFact across multiple backbone LLMs, bold denotes the best performance, while underline denotes the second-highest performance. } 
\label{tab:main}
\end{table*}

\subsection{Extractor}

Extractor aims to produce both supporting evidence $E$ and candidate answers $A$ for the original question $Q$.
Given the rewrite statement $S$ and the counterfactual statement $R^{*}$, the model is prompted to perform step-by-step reasoning grounded in the table $T$\footnote{We implement our method within the Table-Critic framework for execution convenience, but our approach can be deployed within other table reasoning systems as well.}, resulting in two reasoning traces, $\mathrm{Trace}(S)$ and $\mathrm{Trace}(R^{*})$.
Each reasoning trace is treated as an instantiation of the corresponding latent reasoning space $\Psi(X)$.

We collect evidence from each trace as follows:
\begin{equation}
\begin{aligned}
\mathbb{I}_{T,Q}(e) &=
\begin{cases}
1, & \text{if } e \text{ is consistent w.r.t.\ } (T,Q),\\
0, & \text{otherwise},
\end{cases} \\[4pt]
E(X) &= \sum_{e \in \mathrm{Trace}(X)} \mathbb{I}_{T,Q}(e),
\qquad X \in \{S, R^{*}\}.
\end{aligned}
\end{equation}
Here, $\mathbb{I}_{T,Q}(e)$ using an LLM to evaluate whether a candidate evidence item $e$ appearing in $\mathrm{Trace}(X)$ is logically consistent under the table--question context $(T,Q)$ and informative for answering the question.
Items that fail this check are discarded, while those with $\mathbb{I}_{T,Q}(e)=1$ are accumulated along the trace to form the evidence associated with $X$, denoted $E(X)$.

Applying this procedure to the two traces yields two evidence sets, $E_{1}=E(S)$ and $E_{2}=E(R^{*})$, corresponding to the Rewriter and Reverser reasoning processes, respectively.
Each evidence set is subsequently used together with the original question and table to produce a candidate answer via the same large language model, i.e., $A_{1}=\mathcal{M}(Q,T,E_{1})$ and $A_{2}=\mathcal{M}(Q,T,E_{2})$.

\subsection{Rethinker}

We design a set of weighted decision rules for the Rethink module. When $A_{1}$ and $A_{2}$ coincide, the module directly returns the shared answer. When the two candidate answers differ, we compute self-consistency scores for each $(\text{answer}, \text{evidence})$ pair produced by Rewriter and Reverser, mapping each score into the range $[-1, 1]$. The module then compares the score difference and determines the answer according to a predefined set of rules. For cases in which the two candidates cannot be reliably distinguished, the model re-evaluates each candidate under the opposing evidence to test whether its reasoning remains consistent under counter perspectives. The detailed decision algorithm is provided in Appendix~\ref{app:rethink}.

\section{Experiments}

\subsection{Experimental Setup}

\paragraph{Datasets} We evaluate the proposed CRAFT on two widely used table-understanding
datasets: TabFact \citep{chen2020tabfactlargescaledatasettablebased} and WikiTQ \citep{Pasupat2015}

TabFact is a table-based binary fact verification (FV) dataset in which the task is to determine whether a given statement is supported by the table. Following prior work, we report binary classification accuracy as the evaluation metric. In contrast, WikiTQ is a table-based question answering (QA) dataset with short, structured answers. We evaluate performance using the official denotation accuracy.

For all datasets, we conduct systematic evaluations using several strong base models, including Deepseek-R1-distilled-14B\citep{Guo_2025}, LLaMA 3.3--70B \citep{Llama32024}, Qwen 2.5--Instruct \citep{qwen2025qwen25technicalreport}, and GPT-5-mini \citep{openai2025gpt5systemcard}.\footnote{For each model family, we report the best-performing version observed within the models we evaluated. } The full details of all tested model settings, prompt formats, hyperparameters, and additional experiments settings are provided in Appendix~\ref{sec:appendix-implement}.

\paragraph{Baselines} We compare CRAFT with three representative categories of baseline methods:

\textbf{(1) Standard reasoning} methods typically rely on direct prompting. End-to-End QA feeds the table and the question directly into the model, without any explicit intermediate reasoning.
Few-Shot QA extends this setting by conditioning the model on a small set of in-context table–question–answer examples, from which the model implicitly learns the desired input–output patterns and reasoning regularities.

\textbf{(2) Task decomposition} methods decompose table reasoning into structured subtasks. Binder \citep{cheng2023binding} translates questions into executable SQL or Python programs, enabling interpretable symbolic reasoning. 
Dater \citep{ye2023versatiledecomposers} follows a parsing–execution–filling process and performs question decomposition at the sub-table level. 
Chain-of-Table \citep{wang2024chainoftable} incrementally constructs sub-tables, with each step building on the previous one. ALTER \citep{zhang2025alter} retrieves a relevant subset of data and augments it with schema and semantic information.

\textbf{(3) Critic-based} methods refine reasoning through self-evaluation or multi-agent critique.
Critic-CoT \citep{zheng2024criticcot} applies a self-critique stage that reviews and refines generated CoTs to revise erroneous inference. 
Table-Critic \citep{yu2025tablecritic}  adopts an agentic multi-stage workflow that iteratively evaluates, critiques, and revises intermediate reasoning steps to enhance logical consistency and reliability.

\subsection{Main Results}

Table~\ref{tab:main} reports the performance of our method and representative baselines under different backbone LLMs.

Across the four backbone LLMs, our method achieves an average of 82.4\% on WikiTQ and 94.6\% on TabFact, exceeding the strongest baseline by 4.7 and 1.1 points, respectively. This improvement is observed for all four backbones. Notably,  the largest absolute gain occurs with Llama3.3--70B, where WikiTQ accuracy increases by 11.3 points over the strongest baseline. In standard baseline settings and at comparable model scales, Qwen2.5-72B consistently outperforms Llama3.3-70B. With our framework in place, however, this ordering is substantially altered: Llama3.3-70B surpasses Qwen2.5-72B on WikiTQ and TabFact. This reversal indicates that our method significantly reduces cross-model performance discrepancies.

Beyond absolute performance, the results reveal two systematic patterns. First, although open-ended TableQA remains more challenging than fact verification, our framework yields larger relative gains on WikiTQ, providing direct evidence that casting QA into a fact-verification--style reasoning paradigm is an effective modeling choice.
Second, our method exhibits uniform applicability across a wide range of model regimes, spanning different model families, different parameter scales, and both open-source and closed-source LLMs. This robustness further extends to different reasoning instantiations: even when executed by a simpler table-reasoning framework such as Chain-of-Table, our method still outperforms other baseline methods, suggesting that it serves as a general reasoning scaffold rather than backbone-dependent optimization.

Taken together, these results indicate that counterfactual reasoning functions as a general reasoning heuristic rather than a task-specific solution. By guiding how models explore alternatives, it provides a more effective way of thinking about table reasoning, leading to consistent performance gains across settings.

\subsection{Ablation Study on Rewriter and Reverser}
As shown in Table~\ref{tab:rewrite-reverse-backbone}, we conduct ablation experiments to verify the effectiveness of Rewriter, Reverser and their combination. For TabFact, where inputs are already in declarative form. Taking LLaMA3.3--70B as an example, Rewriter-only matches the baseline performance of 92.1\%, Reverser-only follows the counterfactual path alone and drops slightly to 91.9\%. This difference is as anticipated because counterfactual reasoning deliberately explores alternative and potentially incorrect hypotheses. Notably, the comparable accuracy between the two paths indicates that solving counterfactual statements can still recover sufficient evidence to answer the original question.

In TableQA, however, Rewriter-only leads to a substantial improvement, highlighting the benefit of converting questions into declarative statements. The additional gain from combining Rewriter and Reverser suggests a complementary interaction, where counterfactual reasoning broadens the evidence considered for the original question.

\begin{table}[t]
\captionsetup[table]{position=bottom}
\centering
\small
\setlength{\tabcolsep}{5pt}
\renewcommand{\arraystretch}{1.15}
\begin{tabular}{lcc}
\toprule
Method & WikiTQ (\%) & TabFact (\%) \\
\midrule
\multicolumn{3}{l}{\textbf{Qwen2.5-72B}} \\
\midrule
Table-Critic & 77.7 & 92.9 \\
Rewriter only$^{\dagger}$ & 79.1 & 92.9 \\
Reverser only & 78.8 & 91.3 \\
\textbf{CRAFT} & \textbf{80.4} & \textbf{93.9} \\
\midrule
\multicolumn{3}{l}{\textbf{LLaMA3.3-70B}} \\
\midrule
Table-Critic  & 70.1 & 92.1 \\
Rewriter only$^{\dagger}$ & 79.9 & 92.1 \\
Reverser only & 79.7 & 91.9 \\
\textbf{CRAFT} & \textbf{81.4} & \textbf{94.1} \\
\bottomrule
\end{tabular}
\caption{Accuracy of Rewriter-only, Reverser-only, and the CRAFT based on both Rewriter and Reverser across different backbone models. 
$^{\dagger}$ Rewriter differs from the Baseline for WikiTQ but is equivalent for TabFact.}
\label{tab:rewrite-reverse-backbone}
\end{table}

\label{sec:ablation}

\begin{figure}[t]
  \centering
  \setlength{\abovecaptionskip}{2pt}
  \setlength{\belowcaptionskip}{0pt}

  \includegraphics[width=\columnwidth,height=0.22\textheight,keepaspectratio]{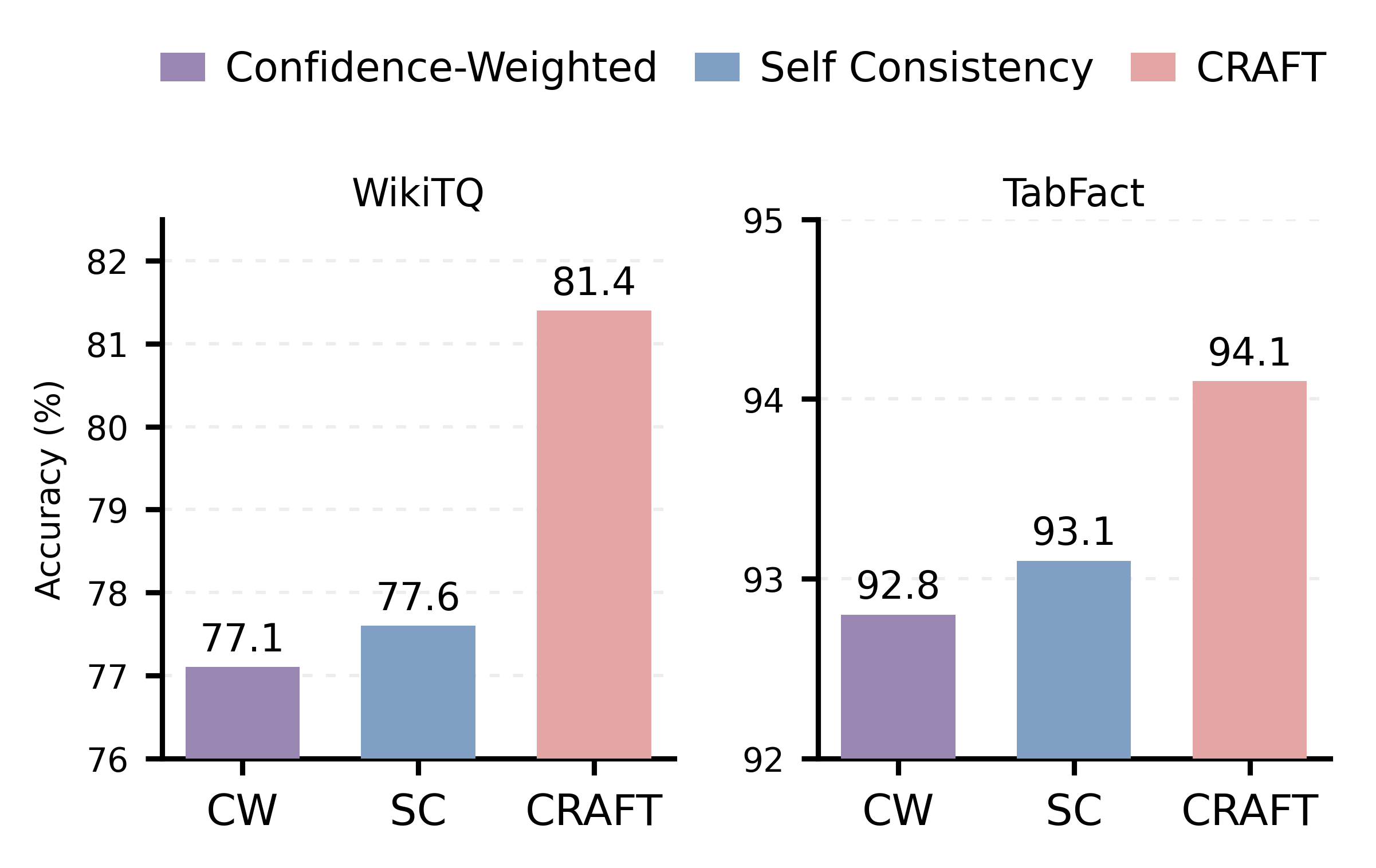}\\[-2pt]
  {\small (a)}

  \includegraphics[width=\columnwidth,height=0.22\textheight,keepaspectratio]{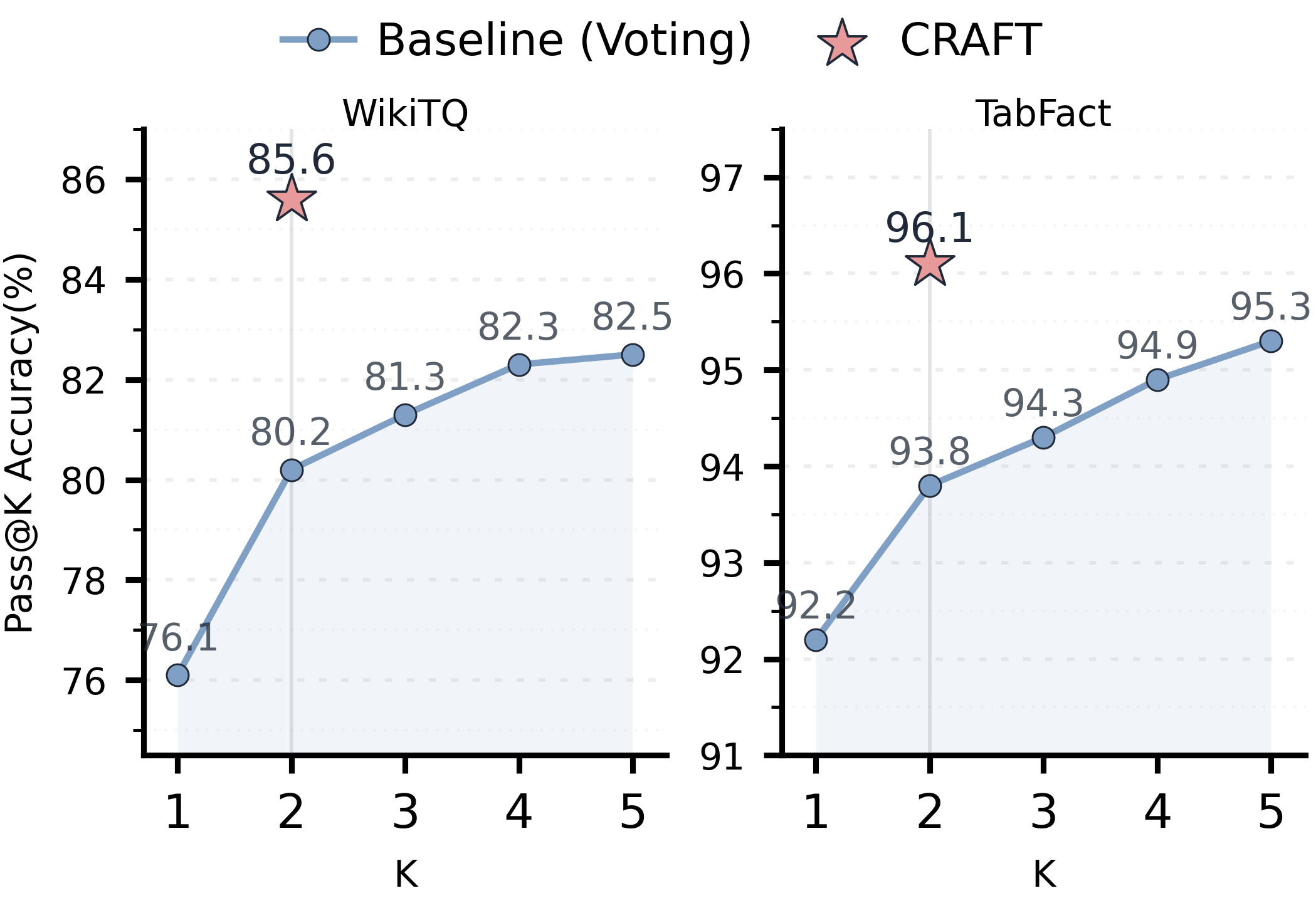}
  {\small (b)}

  \caption{\textbf{Repeated-Sampling analyses on WikiTQ/TabFact.}
  \textbf{(a)} Accuracy comparison between voting methods and CRAFT.
  \textbf{(b)} Ideal(Pass@K) accuracy upper bounds compared with our method.}
  \label{fig:ablation_ab}
\end{figure}

\subsection{Effectiveness Beyond Repeated-Sampling}

Given that our framework involves one more reasoning path in structure, we perform controlled comparisons to eliminate the concern that the observed gains are merely due to repeated sampling. Specifically, we compare the voting method based on Table-Critic with (1) \textbf{Self-Consistency (SC)}, which samples \(N\) answers for each question and selects the final prediction by majority voting over their frequencies; and (2) \textbf{Confidence-Weighted (CW)}, which also samples \(N\) answers but aggregates them by summing \(\exp(\text{score})\) for each unique candidate and choosing the answer with the highest total weight. Here we choose \(N=3\) to allow a fair comparison within a reasonable computational cost.\footnotemark
\footnotetext{From this section onward, all experiments use Llama3--70B unless stated otherwise. Outputs are normalized and format mismatches are not counted as errors, unlike in the main table.}

As shown in Fig.~\ref{fig:ablation_ab}(a), on both table question answering (QA) and table fact verification (FV) tasks, our framework, respectively, achieves improvements of 3.8 and 1.0 percentage points over the best of two voting methods. This indicates that the improvements do not simply stem from ensembling a diverse set of answers.

To further quantify the potential benefit of our method, we adopt the \emph{Pass@K} metric~\citep{chen2021evaluating} to measure whether the correct answer appears among a set of $K$ generated predictions.
Given the set of predictions $\{\hat{y}_1, \dots, \hat{y}_K\}$ and the ground-truth label $y$, Pass@K is defined as:
\begin{equation}
\mathrm{Pass@}K
= \mathbb{I}\!\left(
\exists\, i \in \{1,\dots,K\} \;\text{s.t.}\; \hat{y}_i = y
\right),
\end{equation}
where $\mathbb{I}(\cdot)$ denotes the indicator function.

Pass@$K$ measures whether the candidate answer set contains at least one correct prediction, thereby reflecting how many truly solvable cases are covered by the model’s output distribution and providing an upper bound on the achievable accuracy under $K$ samples\cite{yue2025doesreinforcementlearningreally}.

As reported in Fig.~\ref{fig:ablation_ab}(b), we compare our method with repeated sampling of the strongest baseline under different values of $K$. Our method effectively corresponds to $K=2$, since it explicitly constructs two distinct reasoning paths. Under the same setting, our method achieves a higher Pass@$K$ than the repeated-sampling baseline. Even when the baselines are allowed to increase $K$, they do not match the performance of our approach. This suggests that the direction of sampling—i.e., encouraging diverse and structured exploration of the reasoning space—is more important than the sheer number of samples. By guiding the model to explore different regions of table semantics rather than repeatedly sampling from the same distribution, our method achieves higher coverage of correct reasoning trajectories with only a small number of candidates.

\begin{figure}[t]
  \centering
  \setlength{\abovecaptionskip}{2pt}
  \setlength{\belowcaptionskip}{2pt}

  \includegraphics[width=\columnwidth,height=0.35\textheight,keepaspectratio]{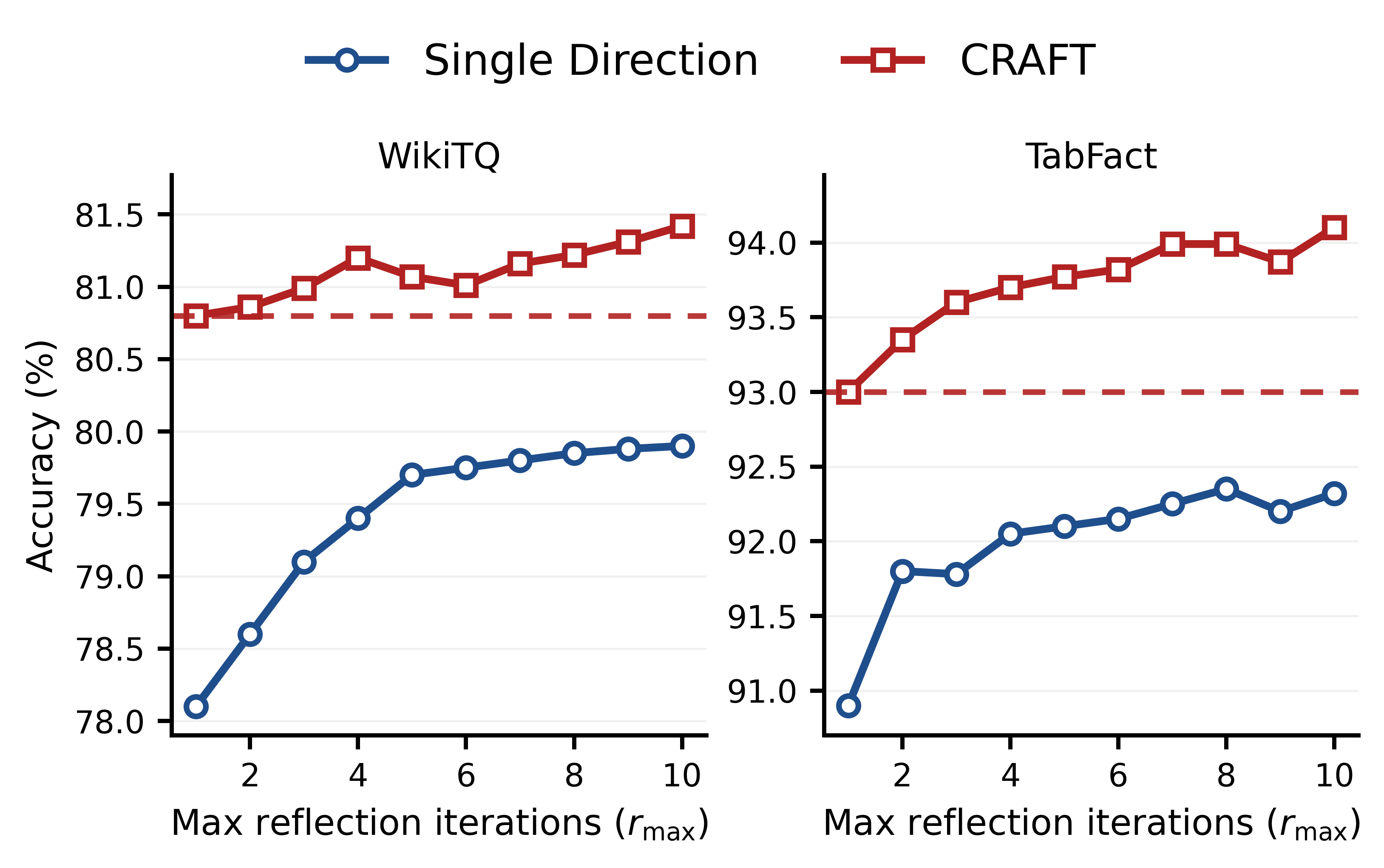}

  \caption{WikiTQ and TabFact accuracy as the number of self-critic iterations increases, comparing CRAFT with single-direction reasoning.}
  \label{fig:ablation_c}
\end{figure}

\subsection{Impact of Self-Critique Iterations} 
The relationship between our method and self-critique is examined by analyzing the effect of varying the number of self-critic iterations. In our method, self-critique is used in the Extractor.
Figure~\ref{fig:ablation_c} shows the effect of varying the number of self-critic iterations on task performance. Across both WikiTQ and TabFact, increasing the number of self-critic iterations consistently improves accuracy compared to the no-self-critic setting, while the performance gain of single-direction reasoning quickly saturate.

Notably, even with an increase in the maximum number of self-critic iterations ($r_{\max}\in[1,10]$) , the performance of single-direction reasoning remains below that of our method, which integrates bidirectional evidences from Rewriter and Reverser. At the same time, the performance of CRAFT also improves as $r_{\max}$ increases, indicating that it is compatible with self-critique. These trends suggest that self-critic effectively strengthens individual candidates, while bidirectional reasoning provides additional, complementary gains that cannot be recovered by increasing self-critic iterations alone.

\subsection{Effect of Multiple Counterfactual Statements} 


To examine whether multiple counterfactual directions can further improve reasoning coverage beyond the single-counterfactual setting, we generate $K$ counterfactual statements ($K \in {1,2,3}$) for each original statement, run the same reasoning pipeline independently on each of them, and evaluate the resulting predictions with Pass@K. As shown in Table~\ref{tab:multi-counterfactual}, increasing $K$ yields consistent but modest gains, with diminishing returns as more counterfactual statements are added. Notably, compared with the repeated-sampling results in Fig.~\ref{fig:ablation_ab}(b), introducing additional counterfactual-driven reasoning paths leads to larger Pass@K improvements than simply repeating the reasoning process. However, this comes at higher computational cost, and although higher Pass@K suggests greater potential benefit from multiple paths, reliably selecting the best one remains non-trivial, which we left for future work.

\begin{table}[t]
\centering
\small
\begin{tabular}{c @{\hspace{10pt}} c @{\hspace{26pt}} c}
\toprule
\multirow{2}{*}{\textbf{\# of Reverse}} & \multicolumn{2}{c}{\textbf{Pass@K (\%)}} \\
\cmidrule(lr){2-3}
& \textbf{WikiTQ} & \textbf{TabFact} \\
\midrule
0 & 79.9 & 92.1 \\
1 & 85.6 & 96.1 \\
2 & 87.7 & 96.5 \\
3 & 88.8 & 96.7 \\
\bottomrule
\end{tabular}
\caption{Pass@K results when using Rewriter with $K$ Reverse (counterfactual) reasoning paths.}
\label{tab:multi-counterfactual}
\end{table}

\subsection{Performance Across Different Table Sizes}

\begin{figure}[t]
  \centering
  \setlength{\abovecaptionskip}{2pt}
  \setlength{\belowcaptionskip}{0pt}

  \includegraphics[width=\columnwidth,height=0.28\textheight,keepaspectratio]{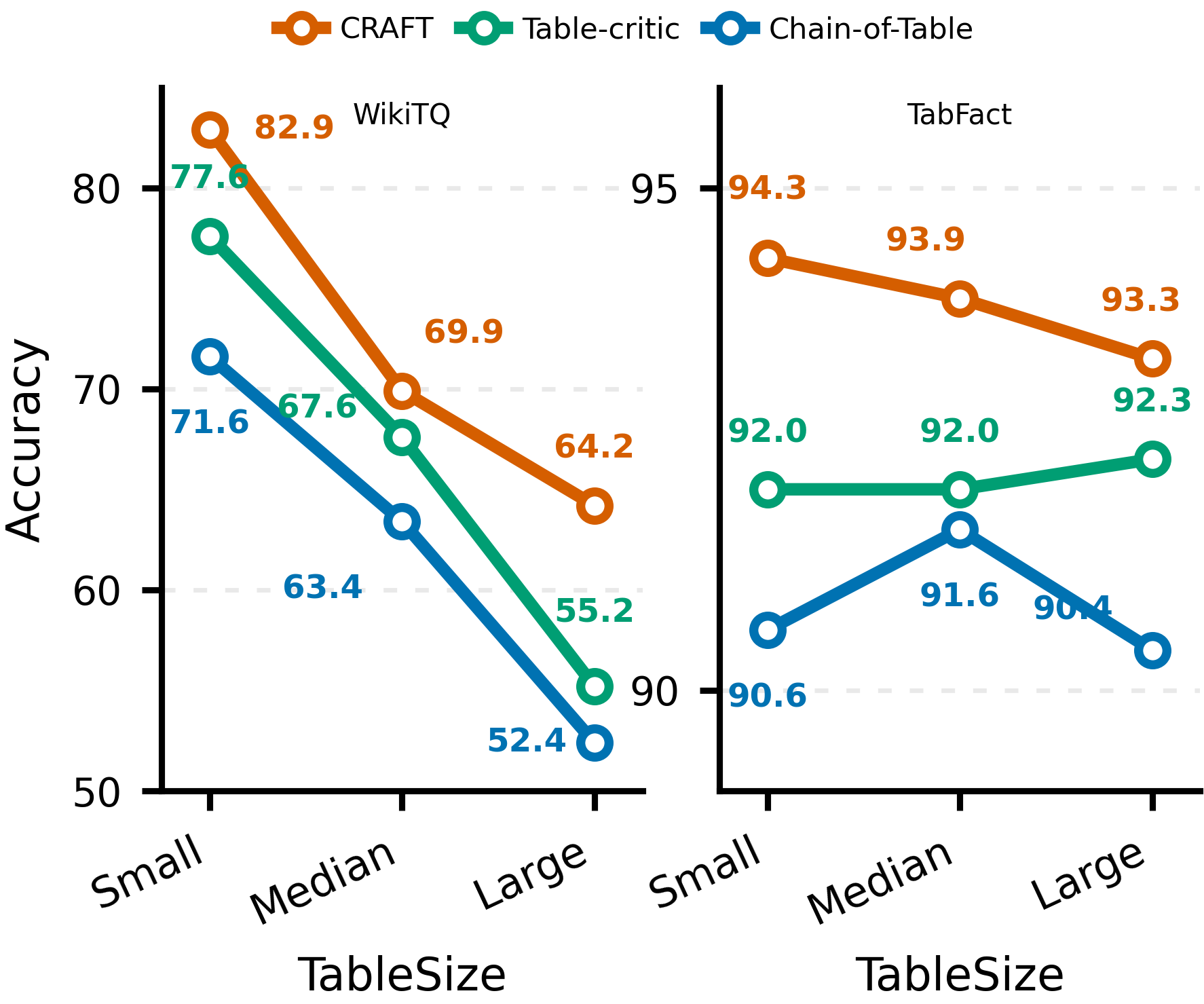}

  \caption{\textbf{Performance across different table sizes.}
  We partition tables into three size groups with thresholds: For WikiTQ, \textit{small} (<2000 tokens), \textit{medium} (2000--4000), and \textit{large} (>4000);
  For TabFact, \textit{small} (<500 tokens), \textit{medium} (500--800), and \textit{large} (>800).}
  \label{fig:ablation_d}
\end{figure}

Large tables pose substantial challenges for large language models (LLMs), as they often struggle to effectively track, integrate, and reason over long input contexts \citep{liu2022tabular, ye2023versatiledecomposers}. To evaluate the impact of table size on performance, we compare our method against two representative baselines, as shown in Figure ~\ref{fig:ablation_d}. 

For WikiTQ, all methods exhibit degraded performance as table size increases, a trend that reflects the growing complexity of reasoning over larger contexts. In contrast, evaluation results of TabFact task show no significant correlation with table size. A plausible explanation is that it primarily requires pinpointing and validating specific claims against relevant table cells and is less sensitive to overall table scale. Notably, our approach demonstrates superior robustness, consistently outperforming baselines across all table sizes. It allows the counterfactual framework to maintain computational efficiency as context length increases.


\section{Conclusion}
We propose CRAFT, an explicit counterfactual framework for table reasoning. The method jointly guides an LLM through both forward and counterfactual reasoning paths, which are integrated via a dedicated rethink module to arrive at the final decision. The framework not only delivers substantial performance gains but also unifies table-based question answering and fact verification under a shared counterfactual reasoning paradigm. Furthermore, the counterfactual perspective provides new insight into how LLMs can be more effectively steered toward structured table understanding.

\section*{Limitations}

While our framework achieves competitive and encouraging results, we acknowledge several limitations that call for continued exploration. First, We did not conduct experiments with smaller base models such as 3B, because tabular inputs typically require long contexts. Second, our method is tailored to table reasoning tasks. Although the underlying ideas may generalize to other forms of reasoning—such as text-based QA—ensuring semantic consistency and generating atomic facts that align closely with ground-truth answers remains highly challenging. Finally, the performance of our \textsc{Rethink} module still leaves room for improvement. While the current results are encouraging, the module has not yet fully realized the potential of counterfactual reasoning. Further refinement is required to maximize its corrective and reasoning capabilities, providing an important direction for future work.

\section*{Ethics Statement}

This work studies a multi-agent reasoning framework for table-based question answering and fact verification. All experiments in this paper are conducted on publicly available benchmark datasets (WikiTQ and TabFact), which do not contain personal or sensitive information. Parts of the implementation and baseline systems are adapted from prior work and publicly released codebases, with appropriate attribution and in compliance with their respective licenses. We view this framework as a research contribution and encourage its use in settings consistent with its intended scope. We therefore believe that this study complies with the ARR Ethics Policy.

\bibliography{custom}
\appendix
\section*{Appendix}
\addcontentsline{toc}{section}{Appendix}

\section{Implementation Details} 
\label{sec:appendix-implement}
All experiments were conducted on a dedicated server equipped with NVIDIA A100 40GB GPUs (×8) with CUDA 12.2. We evaluated our framework using multiple backbone large language models. Open-source models including Llama-3.3-70B and DeepSeek-R1-14B were deployed locally using the vLLM inference engine, while GPT-5-mini was accessed via the Microsoft Azure OpenAI API.

Unless otherwise specified, we used a temperature of 0 and top-$p$ of 1.0. The maximum generation length was set to at most 2048 tokens. In the main experiments, we selected the number of self-critique iterations $r$ from ${1,\dots,10}$ based on validation performance, and used the best-performing value. For API-based models\footnote{We do not report WikiTQ results for \texttt{gpt5-mini} under ALTER because a subset of batches returned empty responses during inference, preventing reliable aggregation of results.}, due to endpoint constraints, we set the temperature to 1. In addition, when inference timeouts occurred, we set the reasoning level to \texttt{minimum} to ensure stable execution, while keeping all other settings unchanged.

\subsection{Output Normalization and Evaluation Metrics} 
In our evaluation, post-processing is applied at two different levels, corresponding to the main results and the supplementary analysis, respectively.
For the main results reported in the primary tables, we apply a strictly minimal normalization only to DeepSeek-R1 distilled models (e.g., DeepSeek-Distilled-14B) by removing explicit reasoning traces or special markers (e.g., \texttt{think} blocks). No matching or answer extraction is performed, and all other models are evaluated on their raw outputs without any post-processing. This setting is intentionally conservative, ensuring that the main comparisons reflect strict adherence to the canonical answer formats of WikiTQ and TabFact.

In the supplementary analysis reported after Section~4.3, we additionally enable a lightweight matching-based extraction on top of the same normalization to better reflect practical usage scenarios in which users accept semantically correct answers even when they are embedded in natural language statements. For example, if a model outputs ``The answer is X'' or similar declarative forms, we extract \texttt{X} and treat it as the predicted answer. Unlike the main results, this matching-based extraction is not restricted to DeepSeek distilled models, and is applied whenever a model's output is non-canonical but contains an unambiguous answer span.
Both stages are conservative and purely surface-level: they neither modify model predictions nor introduce task-specific heuristics, but only standardize syntactic variants for consistent evaluation.

\subsection{TAT-QA Format Alignment and Evaluation Protocol} 
\label{app:tatqa_format_alignment}
For the additional TAT-QA experiments, we do not modify the reasoning procedure or the prompts of any compared baseline. Since TAT-QA is a text-table hybrid dataset, while the evaluated methods are originally designed around table-centric inputs, we only perform input-format alignment to make the dataset compatible with the same evaluation interface. Specifically, we prepend the associated textual context before the table caption in the original table serialization, while keeping the remaining table content and question format unchanged. The same serialized input format is used for all compared methods.  To enable a consistent accuracy-style comparison with WikiTQ, we also canonicalize TAT-QA gold answers into WikiTQ-style answer strings and evaluate predictions with the same WikiTQ evaluator. This conversion only standardizes the surface format of the reference answers, such as answer strings or lists of answer strings, and does not change the semantic content of the gold labels. Therefore, the TAT-QA results reported in the supplementary analysis should be interpreted under this adapted accuracy-style protocol.

\section{Additional Analysis}
\label{app:additional_analysis}

\subsection{Additional Results with Qwen3.5}
\label{app:qwen35_results}

We further conduct supplementary experiments with Qwen3.5 on WikiTQ and TabFact. As shown in Table~\ref{tab:qwen35_results}, CRAFT achieves the best performance on both datasets, with 86.1\% on WikiTQ and 95.3\% on TabFact. This trend is consistent with the main results, indicating that the conclusions remain unchanged when using Qwen3.5-27B as an additional backbone.
\begin{table*}[t]
\centering
\small
\setlength{\tabcolsep}{7pt}
\renewcommand{\arraystretch}{1.10}
\begin{tabular}{lcccccc}
\toprule
\textbf{Dataset} 
& \textbf{Zero-shot} 
& \textbf{Few-shot} 
& \textbf{Chain-of-Table} 
& \textbf{Table-Critic} 
& \textbf{CRAFT + CoT} 
& \textbf{CRAFT} \\
\midrule
WikiTQ  & 70.0 & 71.4 & 80.0 & \underline{85.4} & 85.2 & \textbf{86.1} \\
TabFact & 88.0 & 86.9 & 93.7 & 93.9 & \underline{94.3} & \textbf{95.3} \\
\bottomrule
\end{tabular}
\caption{Performance (\%) of different methods on WikiTQ and TabFact with Qwen3.5 as the backbone.}
\label{tab:qwen35_results}
\end{table*}

\subsection{Evaluation on TAT-QA}
\label{app:tatqa_results}

\begin{table}[t]
\centering
\small
\setlength{\tabcolsep}{7pt}
\renewcommand{\arraystretch}{1.10}
\begin{tabular}{@{}lcc@{}}
\toprule
\textbf{Method} & \textbf{LLaMA3-70B} & \textbf{Qwen2.5-72B} \\
\midrule
Zero-shot & 51.1 & 57.6 \\
Few-shot & 55.9 & 59.2 \\
Chain-of-Table & 56.7 & 55.1 \\
Table-Critic & 61.6 & 65.7 \\
CRAFT + CoT & \underline{68.2} & \underline{72.2} \\
CRAFT & \textbf{71.8} & \textbf{73.9} \\
\bottomrule
\end{tabular}
\caption{Performance (\%) on TAT-QA under the same adapted text-table input setting.}
\label{tab:tatqa_main_results}
\end{table}

We further conduct supplementary experiments on TAT-QA with LLaMA3-70B and Qwen2.5-72B, following the format alignment and evaluation protocol described in Section~\ref{app:tatqa_format_alignment}. As shown in Table~\ref{tab:tatqa_main_results}, CRAFT consistently achieves the best performance under both backbones, reaching 71.8\% and 73.9\%, respectively. This trend is consistent with the main results, indicating that the conclusions remain unchanged in the text-table hybrid setting.

\subsection{Step-0 Candidate Accuracy and Error-Correction Analysis}
\label{app:step0_error_correction}

To examine whether the performance gain mainly comes from the initial candidate already being correct, we conduct an additional WikiTQ analysis with LLaMA3 over three independent runs. Let $y$ denote the gold answer, $\hat{y}_0$ denote the Step-0 candidate, and $\hat{y}$ denote the final prediction. Table~\ref{tab:step0_error_correction} reports the mean Step-0 candidate accuracy and the corresponding error-correction results.

\begin{table}[t]
\centering
\footnotesize
\setlength{\tabcolsep}{4pt}
\renewcommand{\arraystretch}{1.12}
\begin{tabular*}{\columnwidth}{@{\extracolsep{\fill}}llc@{}}
\toprule
\textbf{Metric} & \textbf{Definition} & \textbf{Value} \\
\midrule
Step-0 Acc. & $\Pr(\hat{y}_0 = y)$ & 61.5\% \\
Final Acc. & $\Pr(\hat{y} = y)$ & 81.3\% \\
\hdashline
Correction Rate & $\Pr(\hat{y}=y \mid \hat{y}_0 \neq y)$ & 59.3\% \\
Recovered Share & $\Pr(\hat{y}_0 \neq y \mid \hat{y}=y)$ & 28.1\% \\
\bottomrule
\end{tabular*}
\caption{Step-0 candidate accuracy and error-correction analysis on WikiTQ with LLaMA3, accuracy is averaged over three runs.}
\label{tab:step0_error_correction}
\end{table}

The Step-0 potential-answer accuracy is only 61.5\% on average, substantially lower than the final-answer accuracy of 81.3\%. This indicates that the final performance gain cannot be explained by simply inheriting a correct initial candidate. In particular, CRAFT corrects 59.3\% of initially wrong Step-0 predictions, and 28.1\% of final correct answers come from cases where the Step-0 prediction was wrong. These results show that CRAFT has a strong error-correction ability: even when the initial candidate is incorrect, the reverse path and the Rethinker can often recover the correct answer by integrating complementary table-grounded evidence.

\section{Evaluations on Token Consumption}
\label{sec:appendix-tokens}

\begin{table*}[t]
\centering
\small
\setlength{\tabcolsep}{6pt}
\renewcommand{\arraystretch}{1.15}

\begin{subtable}{0.9\textwidth}
\centering
\begin{tabular*}{\linewidth}{@{\extracolsep{\fill}}lccc@{}}
\toprule
Method & Input Tokens (M) & Output Tokens (M) & Total Weighted Tokens (M) \\
\midrule
Chain-of-Table & 73.5 & 1.6 & 10.6 \\
Table-Critic   & 135.5 & 3.8 & 20.3 \\
CRAFT w/ Chain-of-Table Extractor & 171.2 & 6.2 & 26.8 \\
CRAFT ($c{=}1$) & 234.3 & 8.9 & 37.1 \\
CRAFT ($c{=}5$) & 363.2 & 21.0 & 63.8 \\
\bottomrule
\end{tabular*}
\caption{WikiTQ}
\label{tab:token-consumption-qa}
\end{subtable}

\vspace{0.5em}

\begin{subtable}{0.9\textwidth}
\centering
\begin{tabular*}{\linewidth}{@{\extracolsep{\fill}}lccc@{}}
\toprule
Method & Input Tokens (M) & Output Tokens (M) & Total Weighted Tokens (M) \\
\midrule
Chain-of-Table & 29.3 & 0.6 & 4.2 \\
Table-Critic   & 62.1 & 2.1 & 9.6 \\
CRAFT w/ Chain-of-Table Extractor & 71.1 & 3.4 & 11.9 \\
CRAFT ($c{=}1$) & 104.2 & 4.1 & 16.6 \\
CRAFT ($c{=}5$) & 158.9 & 6.8 & 25.8 \\
\bottomrule
\end{tabular*}
\caption{TabFact}
\label{tab:token-consumption-fv}
\end{subtable}

\caption{Token consumption comparison across different methods on (a) table question answering (WikiTQ) and (b) fact verification (TabFact). Token counts are measured over the full reasoning pipeline, including candidate generation, critic evaluation, and final decision making.}
\label{tab:token-consumption-comparison}
\end{table*}

In this section, we analyze the token consumption of the full reasoning pipeline under different numbers of critic evaluations.
The reported token counts shown in Table~\ref{tab:token-consumption-comparison} cover the entire process, including candidate answer generation, critic-based evaluation, and final decision making.

In our framework, the critic is applied to assess the quality and evidence consistency of candidate answers.
We vary the number of critic evaluations from $c=1$ to $c=10$, where each additional critic evaluation independently reassesses the candidate answers using the same criteria.
As the number of critic evaluations increases, the overall token consumption grows substantially, since each evaluation requires reprocessing the candidate answers together with their associated evidence.

Beyond raw token counts, we also consider a more realistic cost model that reflects current commercial pricing schemes, in which input and output tokens are priced differently.
Following the latest OpenAI pricing, output tokens are substantially more expensive than input tokens; we therefore normalize token usage by weighting output tokens more heavily.
Concretely, we adopt a normalized cost defined as:
\[
\text{Normalized Cost} \;=\; 0.125\,T_{\text{in}} \;+\; 0.875,T_{\text{out}},
\]

where $T_{\text{in}}$ and $T_{\text{out}}$ denote the numbers of input and output tokens, and $P_{\text{in}}$ and $P_{\text{out}}$ denote their respective per-token prices.

As shown in the main results, using a larger number of critic evaluations generally leads to stronger performance, as repeated evaluation helps reduce unreliable decisions.
Therefore, higher values of $c$ are preferred when computational resources permit. At the same time, even a single critic evaluation ($c=1$) already consistently outperforms the baseline, without further critic evaluation, while incurring substantially lower normalized cost. In addition, introducing Extractor stage based on Chain-of-Table, CRAFT significantly reduces token consumption and doesn't introduce critic, while its performance remains among the leading methods in the main results. 

These results indicate a clear trade-off between performance and computational cost controlled by the number of critic evaluations. While multiple critic evaluations yield stronger results, using $c=1$ serves as an economical alternative that retains most of the performance gains with substantially reduced token usage under realistic pricing assumptions.

\section{A Pseudocode Description of Rethinker/Reverser}
This section includes pseudocode and brief supplementary notes on the implementation of the Reverse and Rethink modules.

\begin{table*}[t]
\centering
\small
\setlength{\tabcolsep}{6pt}
\renewcommand{\arraystretch}{1.25}
\caption{Atomic rewriting rules and templates.}
\label{tab:reverse-atomic-rules}
\begin{tabular}{|p{0.10\textwidth}|p{0.32\textwidth}|p{0.54\textwidth}|}
\hline
\textbf{Category} & \textbf{Trigger} & \textbf{Rewriter Template} \\
\hline
Tie
& most / least / highest / lowest / biggest / fastest
& Prefer multiplicity: ``Not only $X$ \dots'' or ``Besides $X$, other entities also \dots''. \\
\hline
Unique
& first / only / sole / unique / earliest / latest / exactly one
& Directly negate the named entity / time / place; do not use ``not only''. \\
\hline
Which
& which + only / solely / uniquely / all
& Use multiplicity: ``Not only $X$ \dots''. \\
\hline
Value
& WHEN / WHERE / HOW MANY / HOW MUCH
& Directly contradict the cited time / place / value, preserving units and context. \\
\hline
Compare
& higher/lower; earlier/later; bigger/smaller
& Flip to the opposite option; never rewrite as Yes/No. \\
\hline
AtLeast
& at least $X$
& Rewrite as: less than $X$. \\
\hline
AtMost
& at most $X$
& Rewrite as: more than $X$. \\
\hline
Exact
& exactly $X$
& Rewrite as: not $X$. \\
\hline
\end{tabular}
\end{table*}

\subsection{The Reverser}
\label{app:reverse}






\begin{algorithm*}[t]
\small
\caption{Reverser Constructor}
\label{alg:reverse}
\begin{algorithmic}[1]
\Require Declarative statement $S$; optional question $Q$; table column names $\mathcal{C}$; reverse template set $\mathcal{D}$
\Ensure Counterstatement statement $R^{*}$

\Statex \textbf{Step 1: Template matching and candidate instantiation} \hfill {\scriptsize$\triangleright$ generate multiple reverse candidates}
\State $\{\delta_1,\dots,\delta_k\} \leftarrow \mathcal{M}_{\textsc{Match}}(S,Q,\mathcal{D})$
\For{$j=1$ to $k$}
    \State $R_j \leftarrow \phi^{\delta_j}(S,Q)$
\EndFor
\State $\mathcal{R} \leftarrow \{R_1,\dots,R_k\}$

\Statex \textbf{Step 2: SQL-signature induction} \hfill {\scriptsize$\triangleright$ use SQL structure as proxy for induced reasoning space}
\State $P_S \leftarrow \mathcal{M}_{\textsc{SQL}}(S,\mathcal{C})$
\State $\Sigma_S \leftarrow \Call{ExtractSignature}{P_S}$
\For{$j=1$ to $k$}
    \State $P_j \leftarrow \mathcal{M}_{\textsc{SQL}}(R_j,\mathcal{C})$
    \State $\Sigma_j \leftarrow \Call{ExtractSignature}{P_j}$
\EndFor
\Statex \hfill {\scriptsize$\triangleright$ each signature is represented by operation--object pairs extracted from minimal SQL}

\Statex \textbf{Step 3: Reverse selection} \hfill {\scriptsize$\triangleright$ maximize structural expansion beyond $S$}
\State $j^{*} \leftarrow \arg\max\limits_{1 \le j \le k} \; \bigl| \Sigma_S \cup \Sigma_j \bigr|$
\State $R^{*} \leftarrow R_{j^{*}}$

\State \Return $R^{*}$
\end{algorithmic}
\end{algorithm*}
The Reverser procedure is summarized in Algorithm~\ref{alg:reverse}. To instantiate reverse candidates, we compile a small set of commonly used reverse templates based on empirical observation of statement semantics. These templates are summarized in Appendix Table~\ref{tab:reverse-atomic-rules} and are used to prompt candidate generation while keeping the candidates semantically aligned with the original statement.

Given the original statement and each candidate reverse statement, we independently prompt the model to generate a minimal SQL-style query for verification. This design is intended to reduce reward hacking during SQL generation: instead of allowing candidates to benefit from arbitrarily long or structurally inflated programs, we explicitly require each statement to be expressed by a minimal verification query.

We first elicit the SQL program itself and then apply rule-based extraction to derive normalized structural representations from the generated SQL. Specifically, we represent each SQL query using operation--object signatures, where each signature associates a SQL operation with its main object. In implementation, reverse selection is based primarily on structural expansion at the operation--object level, namely, the amount of new operation--object structure introduced beyond the original statement. If multiple candidates introduce the same amount of operation--object expansion, we further ask the model to decide which candidate is more helpful for solving the original question. Based on the SQL query patterns and functions discussed by \citep{zhang-etal-2025-high}, we summarizes the SQL operator vocabulary and the corresponding object-level extraction used in our implementation in  Table~\ref{tab:sql-signature-vocab} .

\begin{table*}[t]
\centering
\small
\caption{SQL operator vocabulary and object-level signature construction used in reverse selection.}
\label{tab:sql-signature-vocab}
\begin{tabular}{p{2.2cm}p{4.8cm}p{6.2cm}}
\toprule
\textbf{Operator} & \textbf{Detected from} & \textbf{Object-level signature example} \\
\midrule
\texttt{SELECT}   & \texttt{SELECT ... FROM ...} clause & \texttt{SELECT::height}, \texttt{SELECT::max(height)} \\
\texttt{WHERE}    & \texttt{WHERE} clause & \texttt{WHERE::country = 'Germany'} \\
\texttt{GROUP\_BY} & \texttt{GROUP BY} clause & \texttt{GROUP\_BY::country} \\
\texttt{HAVING}   & \texttt{HAVING} clause & \texttt{HAVING::count(*) > 1} \\
\texttt{ORDER\_BY} & \texttt{ORDER BY} clause & \texttt{ORDER\_BY::height DESC} \\
\texttt{LIMIT}    & \texttt{LIMIT} clause & \texttt{LIMIT::1} \\
\texttt{DISTINCT} & \texttt{DISTINCT} in \texttt{SELECT} expression & \texttt{DISTINCT::distinct country} \\
\texttt{COUNT}    & aggregation call \texttt{COUNT(...)} & \texttt{COUNT::*}, \texttt{COUNT::height} \\
\texttt{SUM}      & aggregation call \texttt{SUM(...)} & \texttt{SUM::sales} \\
\texttt{AVG}      & aggregation call \texttt{AVG(...)} & \texttt{AVG::score} \\
\texttt{MIN}      & aggregation call \texttt{MIN(...)} & \texttt{MIN::year} \\
\texttt{MAX}      & aggregation call \texttt{MAX(...)} & \texttt{MAX::height} \\
\texttt{CASE}     & \texttt{CASE WHEN} or \texttt{IF(...)} expression & \texttt{CASE::case} \\
\texttt{EXISTS}   & \texttt{EXISTS} clause & \texttt{EXISTS::exists} \\
\texttt{SUBQUERY} & nested \texttt{SELECT} query & \texttt{SUBQUERY::subquery} \\
\bottomrule
\end{tabular}
\end{table*}

\subsection{The Rethinker}
\label{app:rethink}
Using large language models as critics or judges has been shown to be effective, but it is also challenging in practice, as their feedback can be noisy, inconsistent, overconfident, or sensitive to prompt phrasing and context \citep{raina-etal-2024-llm,bavaresco-etal-2025-llms,haldar-hockenmaier-2025-rating}. As a result, we do not rely on the model's feedback alone. Instead, based on empirical validation, we adopt a hybrid design that combines model feedback with simple structural constraints. This design allows us to leverage the model's semantic judgments while preventing unstable or unconstrained revisions. To assess the contribution of this design, we report a simple ablation in which the structural constraints are removed and the system relies solely on model feedback (Table~\ref{tab:resync-ablation}). Algorithm ~\ref{alg:rethink-full}  provides the pseudocode description of this module.

\section{Case Study and Prompts}
\label{sec:case-study}

We provide qualitative illustrations and implementation details for each step of the CRAFT. It is organized to support a clearer understanding of how the proposed method operates, how its components interact, and how its behavior differs from that of prior reflection-based approaches.

\begin{table}[t]
\centering
\small
\setlength{\tabcolsep}{5pt}
\renewcommand{\arraystretch}{1.15}
\caption{Ablation study results comparing \textsc{Rethink} with a purely model-based prompting baseline on WikiTQ and TabFact tasks, using Llama 3 as the backbone model.}
\label{tab:resync-ablation}

\begin{tabular}{lcc}
\toprule
Method & WikiTQ Acc. (\%) & TabFact Acc. (\%) \\
\midrule
Prompt-based & 79.9 & 91.7 \\
\textsc{Rethink} & 81.4 & 94.1 \\
\midrule
$\Delta$ & $\uparrow\,{+1.5}$ & $\uparrow\,{+2.5}$ \\
\bottomrule
\end{tabular}%
\end{table}

\begin{figure*}[t]
    \centering
    \includegraphics[width=\textwidth]{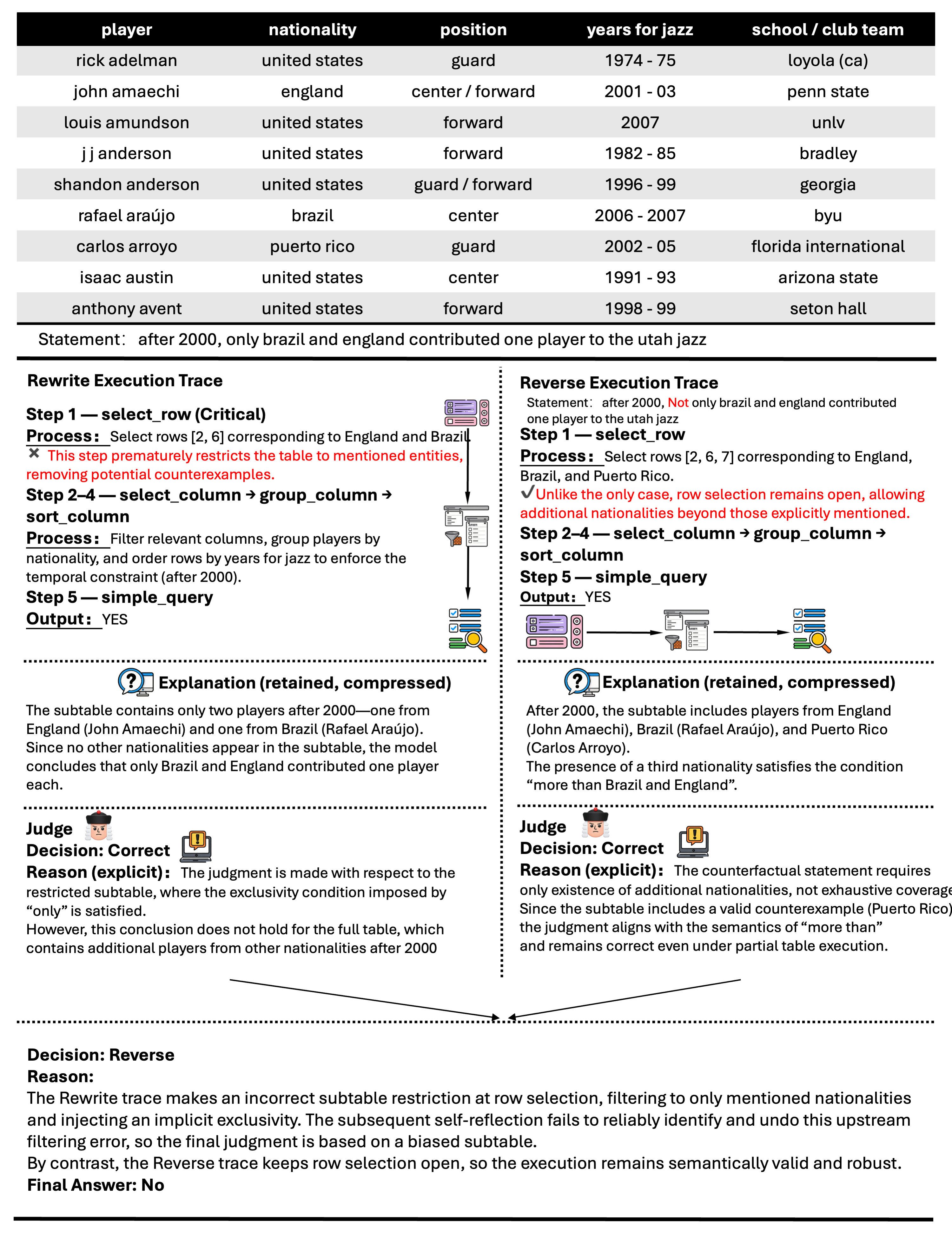}
    \caption{A Case Study Comparing Self-Critique and Counterfactual Reasoning Paths}
    \label{fig:case-study}
\end{figure*}

\begin{figure*}[t]
    \centering
    \includegraphics[width=\textwidth]{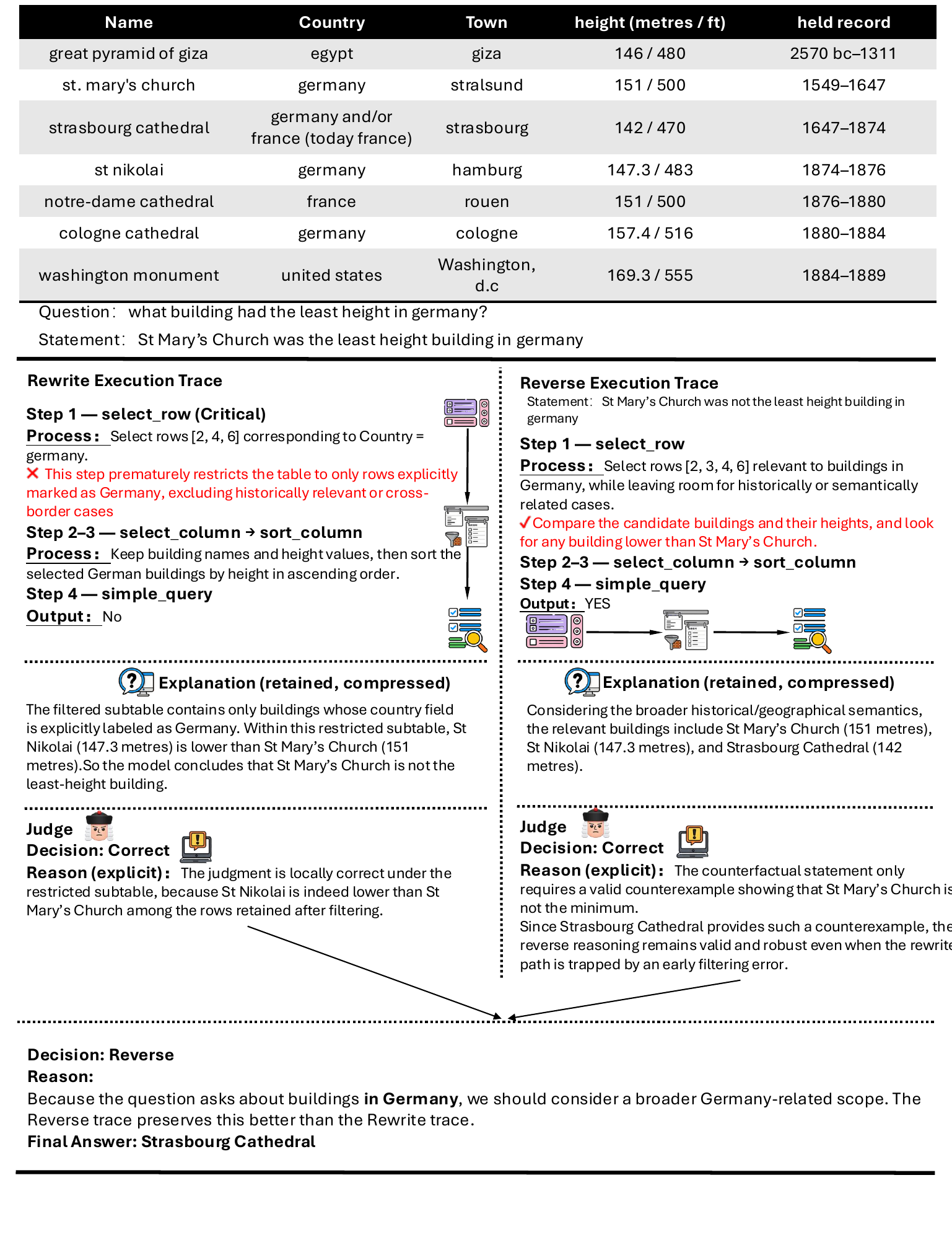}
    \caption{A Case Study Showing how CRAFT get correct answer when Both Reasoning Paths start with a wrong answer}
    \label{fig:case-study2}
\end{figure*}

We first present a detailed case study ~\ref{fig:case-study} that traces the reasoning process of our method and contrasts it with that of a representative self-reflection--based baseline on the same example. By comparing the key intermediate steps under the two settings, the case study illustrates how counterfactual construction leads the model to explore alternative reasoning paths, and how inappropriate intermediate restrictions in prior approaches can result in incorrect conclusions, whereas the proposed framework reaches a consistent and correct outcome. In a separate example~\ref{fig:case-study2} , we further show that CRAFT can still approach the correct solution even when neither the Rewriter nor the Reverser statements include the gold answer. CRAFT leverages the partial but complementary clues uncovered along the two reasoning paths. This allows CRAFT to filter out less reliable candidates, retain the more informative comparisons, and ultimately recover the correct answer through the final evidence aggregation stage.

\begin{figure*}[t]
    \centering
    \includegraphics[width=\textwidth]{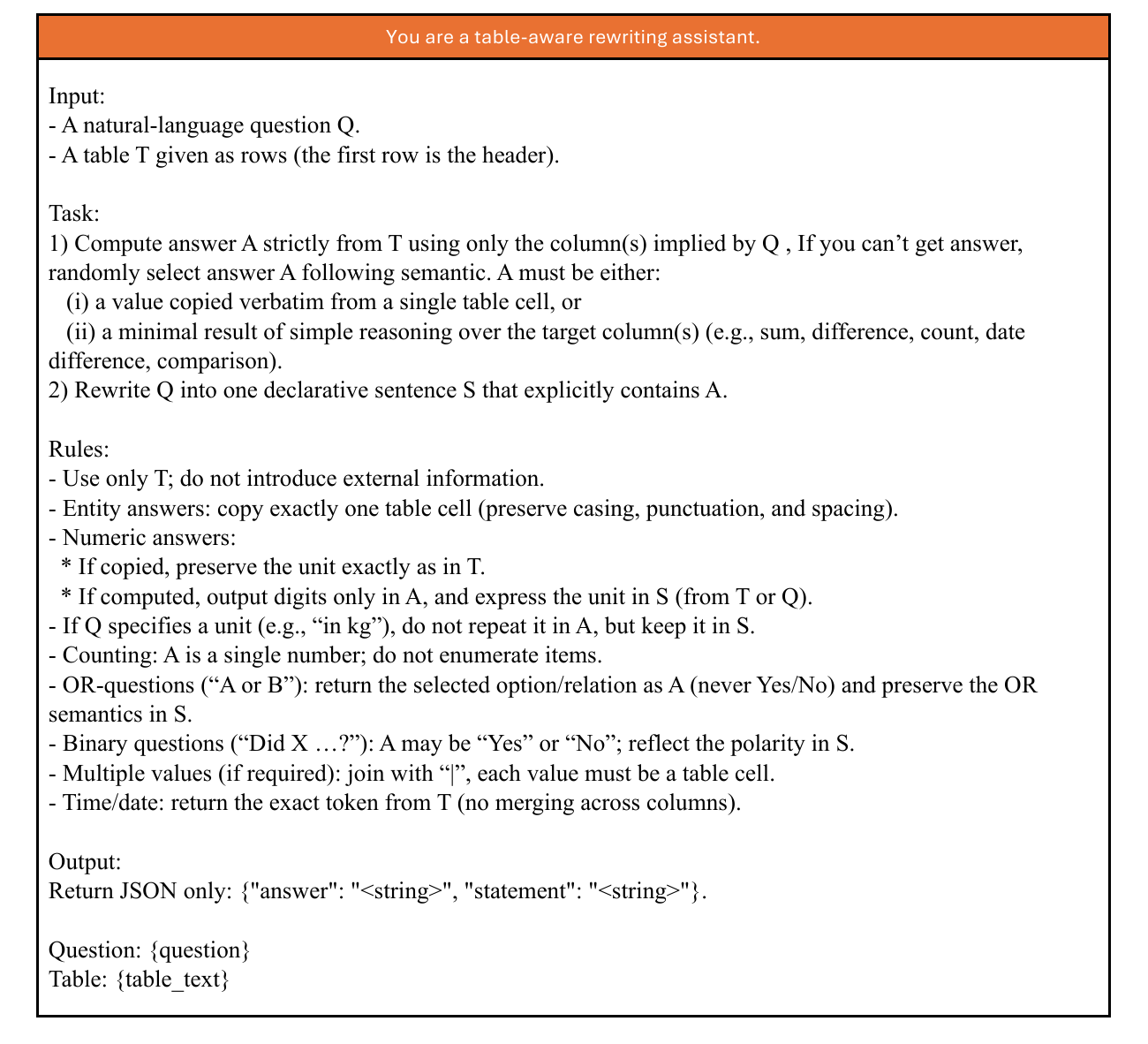}
    \caption{Rewriter Prompt Template.}
    \label{fig:prompt-rewrite}
\end{figure*}

Following this, we report the exact prompts used for all modules in CRAFT to ensure transparency and reproducibility. Specifically, we provide the prompts for the Rewriter, Reverser, Extractor, and Rethinker components (see Figure~\ref{fig:prompt-rewrite}). Since the intermediate reasoning traces produced by baseline methods can be directly used as evidence inputs, we only include the Extractor prompt that maps a given piece of evidence to a final answer. This prompt defines how evidence is interpreted and converted into a prediction.

\begin{algorithm*}[t]
\caption{Rethink Component}
\label{alg:rethink-full}
\normalsize
\begin{algorithmic}[1]
\Require Question $Q$, table $T$; Rewriter $(a_A,e_A)$; Reverser $(a_B,e_B)$; fallback answer $a_{\mathrm{base}}$; thresholds $\tau,\tau_{\mathrm{swap}}$.
\Ensure Selected answer $\hat{a}\in\{a_A,a_B\}$.

\State \textbf{Function} \Call{Score}{$Q,T,a,e$}:
\State $S \gets \Call{ToStatement}{Q,a}$
\State $(y,\alpha) \gets \Call{LLMJudge}{S,e,T}$ \Comment{$y\in\{\mathsf{Support},\mathsf{Contradict}\},\ \alpha\in[0,1]$}
\If{$y=\mathsf{Support}$}
    \State \Return $+\alpha$
\Else
    \State \Return $-\alpha$
\EndIf

\Statex
\Statex\hspace{-\algorithmicindent}\textbf{Main-view margin (self-evidence)}
\State $s_A \gets \Call{Score}{Q,T,a_A,e_A}$,\;\; $s_B \gets \Call{Score}{Q,T,a_B,e_B}$ \Comment{$s\in[-1,1]$}
\State $\Delta \gets |s_A-s_B|$
\If{$\Delta \ge \tau$}
    \State \Return $\arg\max_{a\in\{a_A,a_B\}} \; s(a)$
\EndIf
\State \Comment{If $\Delta<\tau$, Stage 1 is inconclusive and we enter Stage 2}

\Statex
\Statex\hspace{-\algorithmicindent}\textbf{Swapped-evidence}
\State $s_{A\leftrightarrow} \gets \Call{Score}{Q,T,a_A,e_B}$,\;\; $s_{B\leftrightarrow} \gets \Call{Score}{Q,T,a_B,e_A}$
\State $c_A \gets \mathbb{I}\!\left[s_A \cdot s_{A\leftrightarrow} > 0\right]$,\;\; $c_B \gets \mathbb{I}\!\left[s_B \cdot s_{B\leftrightarrow} > 0\right]$
\State \Comment{$c=1$: sign preserved; $c=0$: sign flips}
\State $\Delta_{\mathrm{swap}} \gets |s_{A\leftrightarrow}-s_{B\leftrightarrow}|$

\Statex
\Statex\hspace{-\algorithmicindent}\textbf{Decision}
\If{$c_A \neq c_B$}
    \If{$\Delta_{\mathrm{swap}} \ge \tau_{\mathrm{swap}}$}
        \State \Return $a_{\arg\max_{v\in\{A,B\}} \; c_v}$
    \EndIf
\EndIf
\State \Return $a_{\mathrm{base}}$

\end{algorithmic}
\end{algorithm*}

\clearpage
\clearpage

\begin{figure*}[t]
    \centering
    \includegraphics[width=\textwidth]{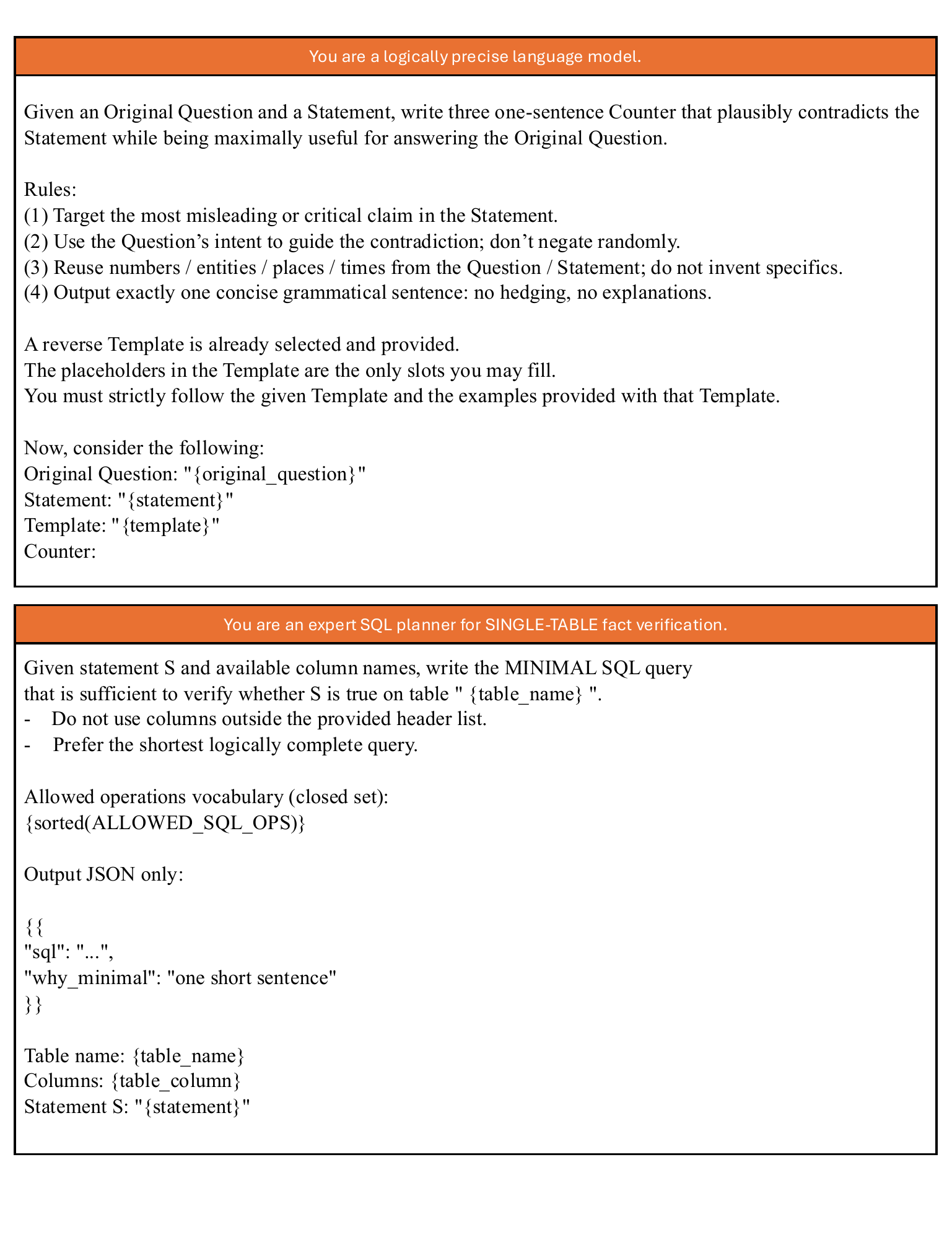}
    \caption{Reverser Prompt Template.}
    \label{fig:prompt-reverse}
\end{figure*}

\begin{figure*}[t]
    \centering
    \includegraphics[width=\textwidth, trim=0 12pt 0 0, clip]{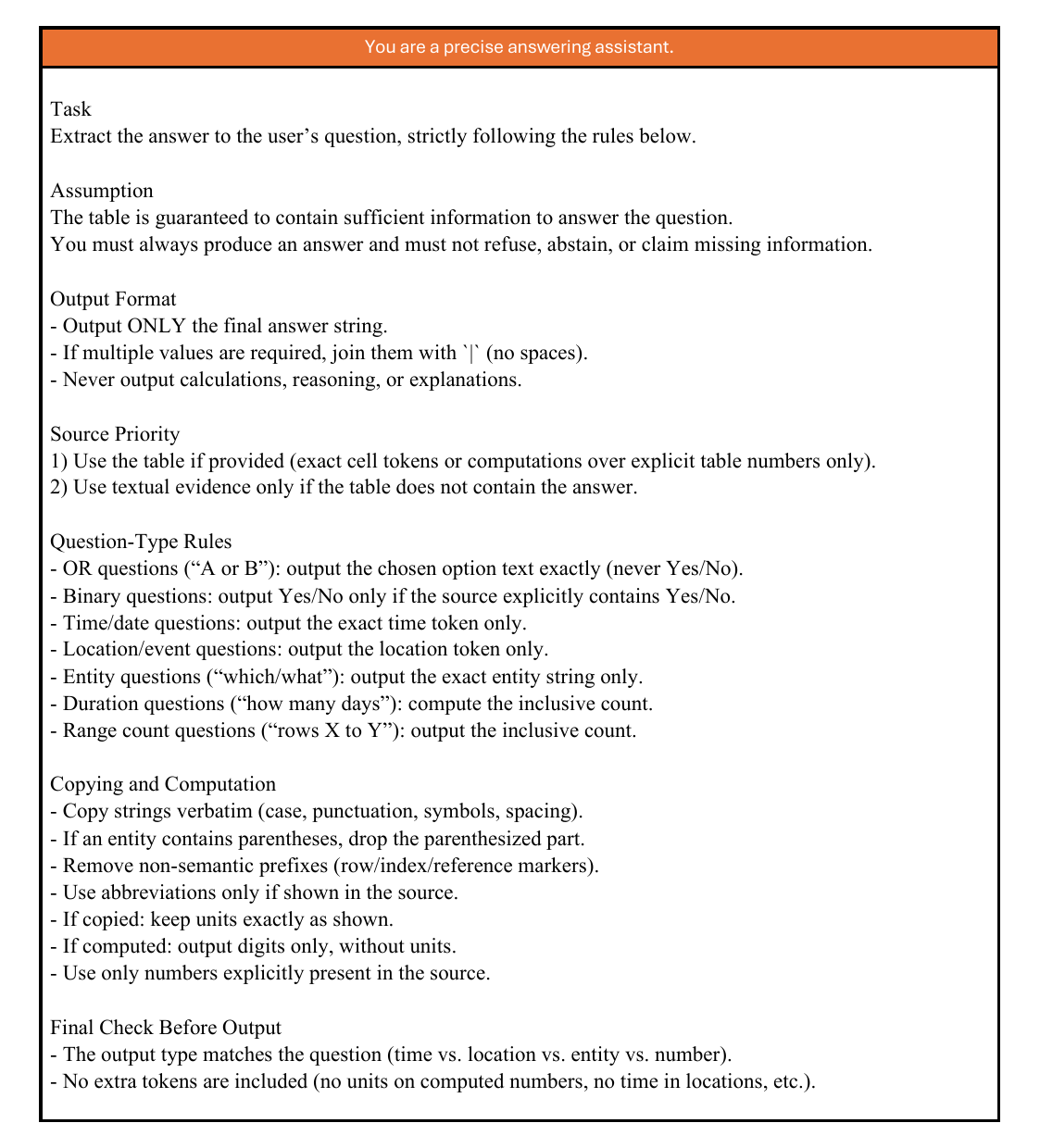}
    \caption{Extractor Prompt Template.}
    \label{fig:prompt-extractor}
\end{figure*}

\begin{figure*}[t]
    \centering
    \includegraphics[width=\textwidth]{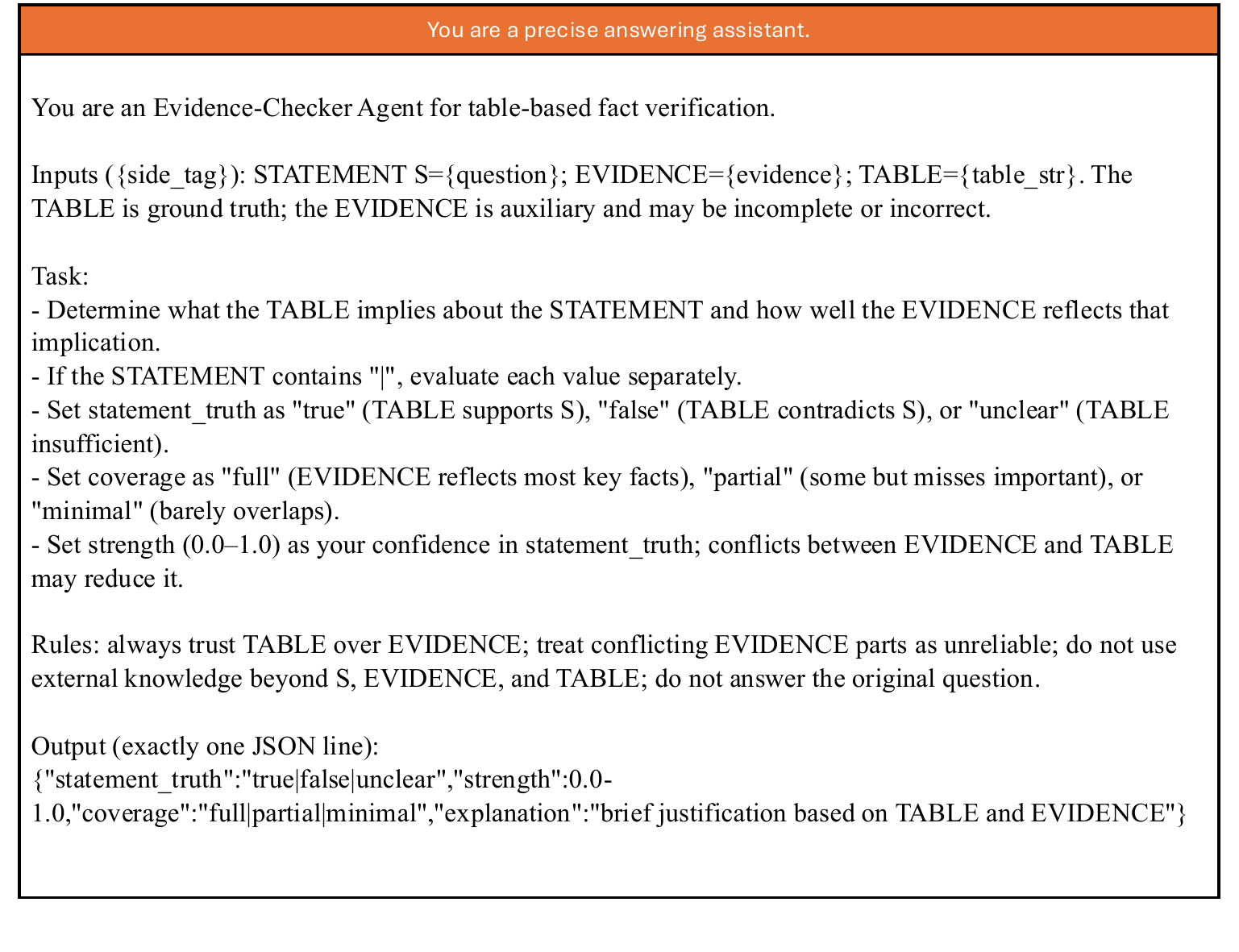}
    \caption{Rethinker Prompt Template.}
    \label{fig:prompt-rethink}
\end{figure*}

\end{document}